\documentclass{article}
\usepackage[utf8]{inputenc}
\usepackage{multirow}
\usepackage{hyperref}

\title{wisardpkg - A library for WiSARD-based models}
\author{Aluízio S. Lima Filho$^1$, Gabriel P. Guarisa$^1$, Leopoldo A. D. Lusquino Filho$^1$,\\ 
Luiz F. R. Oliveira$^1$, Felipe M. G. Fran\c{c}a$^1$, Priscila M. V. Lima$^1$ $^2$\\
$^1$ PESC/COPPE/UFRJ, $^2$ NCE/UFRJ}

\usepackage{graphicx}

\begin{document}
\nocite{*}

\maketitle

\begin{abstract}
    In order to facilitate the production of codes using WiSARD-based models, LabZero developed an ML library C++/Python called wisardpkg. This library is an MIT-licensed open-source package hosted on GitHub under the license.
\end{abstract}

\section{Introduction}

Weightless  artificial  neural  networks  (WANN)  are  neural  models
that  do  not  use weighted synapses to store the information it learns from presented patterns. Alternatively, it possesses RAM (random-access-memory)-based neurons in which information storage takes place. In a WANN, learning of a pattern corresponds to writing in memory, whereas classification essentially corresponds to the reading of certain memory positions. The advantages of these models lie essentially in their speed, simplicity, low computational and power costs and, above all, in their ability to perform online learning.

The WiSARD (Wilkes, Stonham and Aleksander Recognition Device) \cite{aleksander} is a pioneering WANN model that was originally designed to solve simple classification tasks. WiSARD has received extensions to perform semi-supervised and unsupervised learning, regression tasks, and improvements in training and classification policies.

Recent applications of WiSARD with remarkable results corroborating the choice of this model: data stream clustering \cite{douglas, douglas2, douglas3}; time-series classification \cite{souza2014}; audio processing \cite{souza2015}; online tracking of objects \cite{nascimento2015}; GPS trajectory classification \cite{barbosa2018}; part-of-speech tagging \cite{carneiro2015, carneiro2017}; text categorization \cite{rangel2016}; hardware assisted security \cite{santiago2019}; emotional analysis \cite{vidal2013, lusquinofilho2018, aus, empathy}, and; prediction tasks \cite{rew, lusquino2020}. Also, recent studies on theoretical aspects of the model can be found in \cite{carneiro2019}. 

To facilitate the application of WiSARD and its extensions, a library in C/C++ was created, with a wrapper for Python, called wisardpkg. Section 02 of this work describes the models present in wisardpkg, while Section 03 gives details of their implementation, use, configurations and installation.

\section{WiSARD-models in wisardpkg}

\subsection{WiSARD}

WiSARD\cite{aleksander} is a $n$-tuple classifier composed by class \textit{discriminators}; each discriminator is a set of $N$ RAM nodes having $n$ address lines each. All discriminators share a structure called \textit{input retina}, from which a pseudo-random mapping of its $N * n$ bits composes the input address lines of all of its RAM nodes.

\begin{figure}%
    \centering
    \includegraphics[scale=0.3]{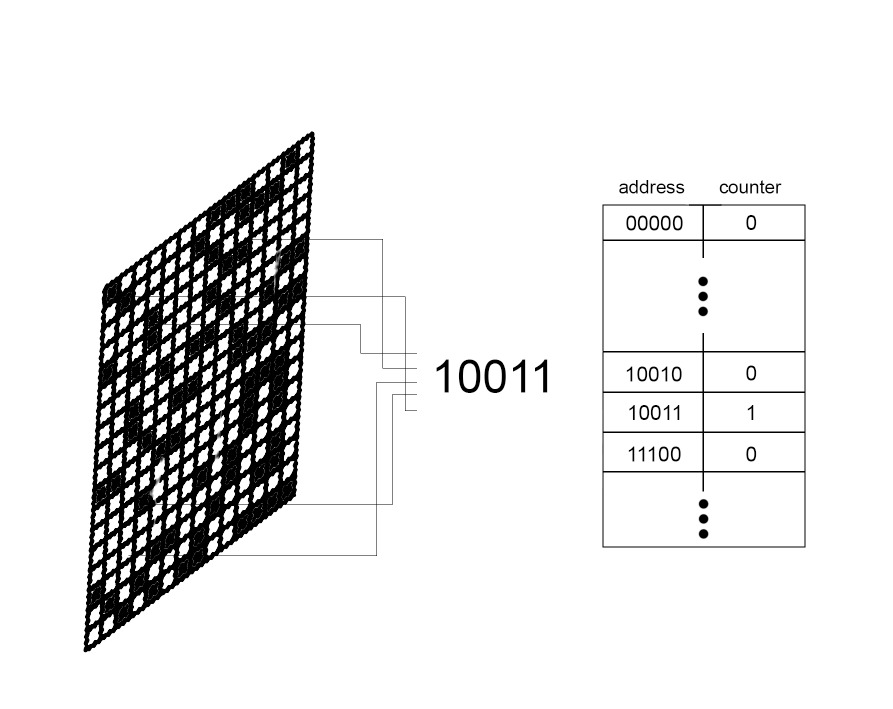}
    \caption{Training in WiSARD\cite{qualify}.}
    \label{fig:wisard}
\end{figure}

WiSARD is initialized with all its memory locations with value “0”. During training, when a binary pattern is presented to the network, it will access the corresponding memory positions in the appropriate discriminator, changing those with null content to “1”. This process is showed in Fig. \ref{fig:wisard}. In the classification phase, the binary pattern accesses all discriminators in the corresponding positions and each of them will return a score formed by the number of non-null positions accessed. The discriminator with the highest score will determine the class of the input.

An extension to the model was made to deal with a learning saturation problem. It consists of replacing the content of the memory positions from a singe bit to an access counter, which is increased during the training phase. In the classification phase, the discriminators' score becomes the sum of the non-null memory positions accessed, whose counter has a value higher than a threshold called \textit{bleaching}\cite{grieco2010}, which is initialized with a value of “0” and is increased whenever there is a tie. In this case, the classification process is repeated until there is a tie and if the value of bleaching becomes greater than the value of the largest access counter, the network will randomly draw the class of one of the discriminators with a score tied to be the class of the input.

\subsection{ClusWiSARD}

ClusWiSARD\cite{douglas} is a variation of WiSARD that allows the same class to have more than one discriminator, so that sub-profiles of the same class, which do not have enough similarity between them, are learned in different places, in order to avoid saturation of the learning of a discriminator with the superposition of extremely heterogeneous patterns, but that still belong to the same class.

ClusWiSARD is initialized with only one discriminator of each class and as new examples are learned, a verification is made to see if there is a need to create a new discriminator. In this model, the same example can be learned in more than one discriminator.

This model can also be used for semi-supervised learning, where when a non-labeled example is submitted for learning, a classification occurs and the discriminator with the highest score will learn the example.

\subsection{Regression WiSARD}

\begin{figure}%
    \centering
    \includegraphics[scale=0.3]{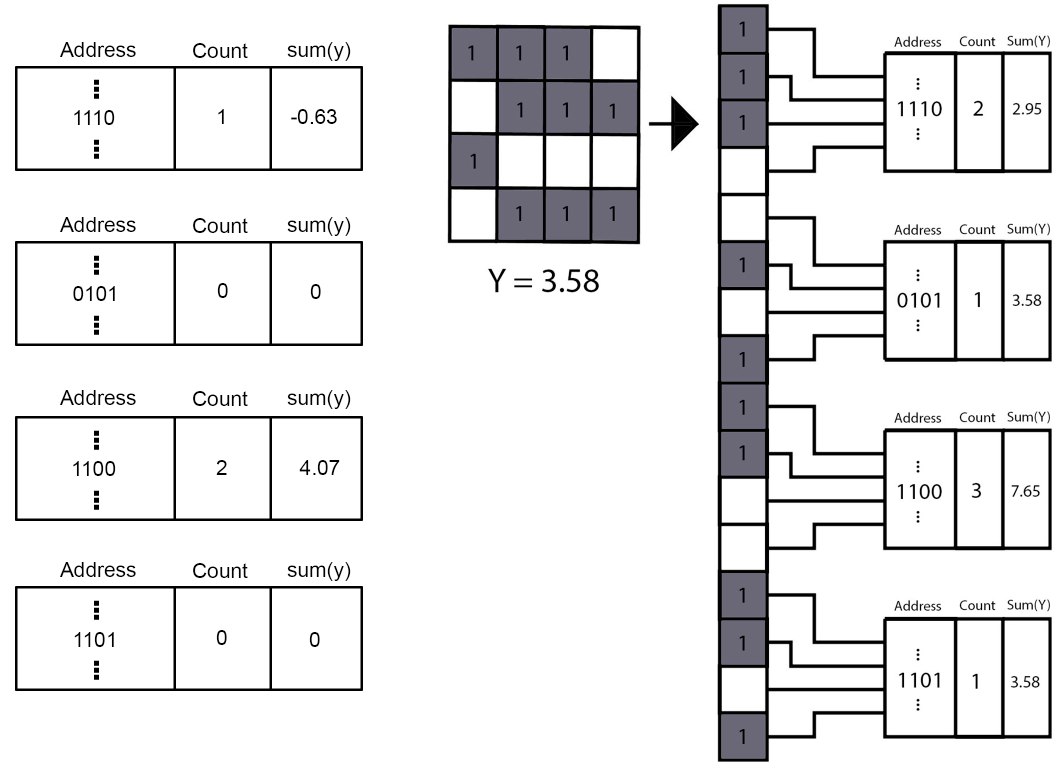}
    \caption{Regression WiSARD prediction\cite{qualify}.}
    \label{fig:rew}
\end{figure}

An extension of WiSARD to handle prediction tasks\cite{rew, lusquino2020}. This model works with just one discriminator with each memory location having two contents/dimensions: (i) an access counter (same as in the WiSARD with bleaching), and; (ii) a partial prediction. During training, when a pair <\textbf{x}, \textit{y}> is presented to the network, \textbf{x} is used to access specific memory locations and increment their access counters. The partial predictions of these same positions are incremented using the value of \textit{y}. At the time of prediction, when an input \textbf{x} is presented to the network, it will access the respective memory positions and Regression WiSARD will return as a prediction an average of the sum of the counters and the partial predictions accessed. This model can use different types of media. This process is showed in Fig. \ref{fig:rew}.

\subsection{ClusRegression WiSARD}

A variation of Regression WiSARD that is based on the same principle as ClusWiSARD of separating examples that are not sufficiently similar\cite{rew, lusquino2020}. This model is initialized with a single ReW discriminator and whenever new examples are learned there is a verification for the need to create new discriminators and a classification to determine which discriminators will learn the example. In the prediction phase, a classification is performed to determine the ReW discriminator that will perform the prediction.

\section{Library's overview}

\subsection{Implementation}

wisardpkg is hosted on GitHub at \hyperlink{https://iazero.github.io/wisardpkg/}{https://iazero.github.io/wisardpkg/}, where users can find the latest version of the library and user documentation. Model details and features described in this publication pertain to the latest version of the model as of the date of this publication. The lib was implemented in C/C++ with a Python wrapper.

\subsection{Availability}

\begin{itemize}
    \item \textbf{Operating systems:} Linux, Mac OSX, Windows
    \item \textbf{Programming languages:} C++ 11 and up, Python version 3.7.0 and up
    \item \textbf{Additional system requirements:} NA
    \item \textbf{Dependency:} pybind11 ($\geq$ 2.5.0)
    \item \textbf{List of contributors:} All contributors were listed as authors with corresponding affiliations
    \item \textbf{Language:} C++, with wrapper to Python 3
    \item \textbf{Current version:} 2.0.0a7
    
\end{itemize}

\subsection{Installation}

\begin{itemize}
    \item \textbf{C++:}
        \begin{itemize}
            \item Clone \hyperlink{https://github.com/IAZero/wisardpkg}{https://github.com/IAZero/wisardpkg}
            \item Copy the wisardpkg.hpp file to the desired project
            \item Include the library in the C++ code
        \end{itemize}
    \item \textbf{Python:}
    \begin{itemize}
        \item Install Python PIP, if necessary
        \item pip install wisardpkg 
    \end{itemize}
\end{itemize}

To install wisardpkg in a Windows environment it is necessary to install Visual C++ additionally. To do this just download it from \hyperlink{https://support.microsoft.com/en-us/help/2977003/the-latest-supported-visual-c-downloads}{here} and then run the installer.

\subsection{Architecture}

The library is divided into two main modules: models and binarization, because since these neural networks only receive binary inputs, it is necessary to treat the input to make it suitable for models. Although this is usually done through some kind of preprocessing external to wisardpkg, the library has some classes to provide support for this pipeline.

\subsubsection{Binarization}

All binarization classes are extensions of the BinBase class. All of them receive an array as input to their unique public method, transform, which will return a binary array. Only the public methods of each class will be described here. \\

\textbf{Thresholding:} applies a simple threshold to a double value to generate a binary input.

\begin{itemize}
    \item Thresholding: its only parameter is the threshold.
    \item transform
\end{itemize}

\textbf{MeanThresholding:} similar to the previous one, but this time the threshold is calculated as the mean of the input data.

\begin{itemize}
    \item MeanThresholding
    \item transform
\end{itemize}

\textbf{Thermometer:} is a technique for preprocessing quantitative variables. Given a variable $d$, a maximum value of traing test $m$ and a number of ranges $s$, the new binary variable will have $s$ bits, with each $i$th bit being determined by a threshold $t = i * \frac{m}{s}$. If $d > t$, the $i$th position is worth $1$, otherwise $0$.

\begin{itemize}
    \item SimpleThermometer: its parameters are the thermometer size, the minimum and the maximum value in its range.
    \item transform
\end{itemize}

\textbf{KernelCanvas:} since each WiSARD-based model is able to handle only one input size, this preprocessing\cite{souza2014} is capable of resizing inputs, being especially useful when dealing with time series. This uses different kernels, or divisions in the sample space of the input, replacing each value of it with the central value of the kernel where it is located.

\begin{itemize}
    \item KernelCanvas: it is possible to instantiate it from a json file. Its parameters are the desired dimensionality and the number of kernels to be used.
    \item transform
\end{itemize}

\subsubsection{Models}

This module contains all the models and also the base classes from which they extend. A brief description of each sub-module and its classes follows.\\

\begin{enumerate}
    \item \textbf{Base:}
    
    \begin{itemize}
    \item \textbf{Model:} a simple trainable module
    \begin{itemize}
        \item train
        \item getsizeof
    \end{itemize}
    \item \textbf{ClassificationModel:} a Model object that can calculate the similarity score in an access, as well as perform classifications
    \begin{itemize}
        \item classify
        \item rank
        \item score
    \end{itemize}
    \item \textbf{RegressionModel:} a Model object that performs predictions. Here the training is overwritten because of the partial prediction used in learning in this type of model.
    \begin{itemize}
        \item train
        \item predict
    \end{itemize}
\end{itemize}

    \item \textbf{Wisard:}
    \begin{itemize}
        \item \textbf{RAM:} the minimal information unit in a weightless neural network, contains $2^{n}$ memory positions. 
        \begin{itemize}
            \item RAM: instantiates RAM. It is possible to use a json file with RAM previously saved for this. Two additional parameters here are ignoreZero (which allows not considering the initial position of each RAM in the classification phase) and base (its default value is 2, forming the classic WiSARD for binary patterns, but when modifying it it is possible to work with patterns that use more bits and the RAM will have $base^{n}$ memory locations).
            \item getVote
            \item train
            \item untrain: it is possible to reverse the training process of an example, once the positions accessed are known, just subtracting their access counters.
            \item getMentalImage: using the retina and the content of the RAM, it generates a representation of the learning.
            \item setMapping: it is possible to choose a mapping for the RAM.
        \end{itemize}
        \item \textbf{Discriminator:}
        \begin{itemize}
            \item Discriminator: its main parameter is the size of the tuple. It is also possible to define your mapping and instantiate it from a json file. Its main methods are: train, untrain and classify.
        \end{itemize}
        \item \textbf{Wisard:} a ClassificationModel that has a set of discriminators. Its main methods are: train, untrain and classify. An optional parameter "balanced" when set to True causes the score of each discriminator during the classification to be normalized using the number of trained examples.
    \end{itemize}
    \item \textbf{Cluswisard:} has only one homonymous class, which will be described bellow:
    \begin{itemize}
        \item Cluswisard: its main parameters are the size of the tuple and the variables used in the verification to create new discriminators: minScore, threshold and discriminatorsLimit.
        \item The main methods here include train, untrain, classify, trainUnsupervised and classifyUnsupervised, the latter is applied only when it is desired to know which is the discriminator with which the example is most similar, despite classes. 
    \end{itemize}
    \item \textbf{RegressionWisard:}
    \begin{itemize}
        \item \textbf{MeanFunctions:} this module has a Mean class, which serves as the basis for several other classes that contain the methods of the means used in the prediction of Regression WiSARD (SimpleMean, PowerMean, Median, HarmonicMean, HarmonicPowerMean, GeometricMean, ExponentialMean and LogisticMean).
        \item \textbf{RegressionRAM:} analogous to the classification RAM, it has an extra content in its memory positions, which is the partial prediction. Additional parameters here include minZero and minOne, which are the minimum amount of these bits that a memory location needs to have to be considered in the prediction phase.
        \item \textbf{RegressionWisard:} The network itself, a set of RegressionRAMs. Its main parameters are the size of the tuple and the average to be used, with minZero, minOne, completeAdressSize and mapping being additional parameters. Like other Models, it can be instantiated from a json file. Its main methods are train and predict.
    \end{itemize}
    \item \textbf{ClusRegressionWisard:} This module has only one homonymous class, which is a RegressionModel, whose main instantiation parameters are addressSize, minScore, threshold and limit. Its main methods are train and predict.
\end{enumerate}

Additionally, the library has a commons module, with exceptions and utils, and a wrapper module for Python.


\begin{footnotesize}

\end{footnotesize}

\end{document}